# Possible World Partition Sequences: A Unifying Framework for Uncertain Reasoning

**Choh Man Teng**
teng@cs.rochester.edu
Department of Computer Science
University of Rochester
Rochester, NY 14627

## Abstract

When we work with information from multiple sources, the formalism each employs to handle uncertainty may not be uniform. In order to be able to combine these knowledge bases of different formats, we need to first establish a common basis for characterizing and evaluating the different formalisms, and provide a semantics for the combined mechanism. A common framework can provide an infrastructure for building an integrated system, and is essential if we are to understand its behavior. We present a unifying framework based on an ordered partition of possible worlds called *partition sequences*, which corresponds to our intuitive notion of biasing towards certain possible scenarios when we are uncertain of the actual situation. We show that some of the existing formalisms, namely, default logic, autoepistemic logic, probabilistic conditioning and thresholding (generalized conditioning), and possibility theory can be incorporated into this general framework.

## 1 INTRODUCTION

Many different formalisms have been proposed for dealing with reasoning under uncertainty. These include default logic [Reiter, 1980], autoepistemic logic [Moore, 1985], circumscription [McCarthy, 1980], probability, belief [Dempster, 1967; Shafer, 1976], and possibility theory [Zadeh, 1978; Dubois and Prade, 1988].

Each formalism has its own idea of how uncertainty in a knowledge base should be handled, and each has provided a different solution. Typically a formalism is characterized using distinct and exclusive syntax and semantics, which are not directly compatible to those of other formalisms, making it difficult to make meaningful comparisons on common terms.

When we work with information from multiple sources, the formats of the knowledge bases and the systems they adopt to express uncertainty may not be uniform. It would be desirable to be able to, for example, combine a knowledge base of default rules with one containing autoepistemic formulas and a third one containing probability assignments. To do so we need to first establish a common basis for characterizing and evaluating the different formalisms, and provide a semantics specifying how the default rules, autoepistemic formulas, and probability statements can be combined and allowed to interact. A common framework can provide an infrastructure for building an integrated system, and is essential if we are to understand and "debug" the behavior of the resulting system.

We propose a unifying framework based on an ordered partition of possible worlds. We call such structures *partition sequences*. A partition sequence corresponds to our intuitive notion of biasing towards certain possible scenarios when we are uncertain of the actual situation. This framework can be adapted to characterize different formalisms by imposing formalism-specific constraints on the way the set of possible worlds can be partitioned. We demonstrate the mechanism by incorporating into the general framework some of the existing formalisms, namely, default logic, autoepistemic logic, probabilistic conditioning and thresholding (generalized conditioning), and possibility theory.

The rest of this paper is organized as follows. Section 2 presents the basic framework of possible world partition sequence. Section 3 gives a brief summary of default logic, autoepistemic logic, probabilistic conditioning and thresholding, and possibility theory. Section 4 gives the details on how the framework can be adapted to these formalisms. Section 5 concludes the discussion. Proof sketches of the theorems can be found in the Appendix.

## 2 POSSIBLE WORLD PARTITION SEQUENCES

The basic structure of our framework is based on possible worlds and an ordering we call a *partition sequence* placed on sets of possible worlds. In a possible world structure, each world corresponds to a possible scenario of the actual world. There is only one real world, but we do not have enough information to determine exactly which world it is. Thus, we have a set of worlds each of which satisfies all the constraints and knowledge we have of the real world.

Although we cannot rule out for sure any of these worlds that are consistent with the current information, there is sometimes a bias on the worlds so that some worlds are considered a more "suitable" model of the real world than others. The measure of "suitableness" of a possible world varies from formalism to formalism. Defaults, probability, and possibility are some of the measures that have been proposed. Many of the formalisms provide some justifications as to how the bias is arrived at, but for our purposes here, it suffices to note each bias satisfies certain constraints (such as the three axioms for probability) which we need to capture in order to characterize the formalism.

We can partition the worlds and order the resulting classes by considering their biases. Worlds that are suitable to the same extent are grouped into the same class, and its place in the partition sequence is determined by the bias of the worlds in the class relative to those in other classes.

## 2.1 NOTATION

Let $\mathcal{L}$ be a standard propositional language, $\mathcal{ML}$ be a standard propositional modal language, and $\mathcal{P}$ be the finite set of propositional constants in $\mathcal{L}$ and $\mathcal{ML}$. We denote the (non-modal) provability operator by $\vdash$. For any set of well formed formulas $S \subseteq \mathcal{L}$, we denote by $\mathbf{Th}(S)$ the set of well formed formulas provable from $S$ by propositional logic; that is, $\mathbf{Th}(S) = \{\phi : S \vdash \phi\}$. Let $\bot$ and $\top$ be the contradiction and tautology symbols.

**Definition 1** *Given a set of elements $W$, a partition sequence of $W$ is a tuple $\langle W_0, \ldots, W_l \rangle$, $l \geq 1$, such that the elements $\{W_i : W_i \neq \emptyset\}$ forms a partition[1] of $W$.*

**Definition 2** *A possible world partition sequence $P$ is a triple $\langle \langle W_0, \ldots, W_l \rangle, m, f \rangle$, where [1] $W = \bigcup_i W_i$ is an exhaustive set of possible worlds, each of which corresponds to a different interpretation of the propositional constants in $\mathcal{P}$. The tuple $\langle W_0, \ldots, W_l \rangle$ constitutes a partition sequence of $W$. [2] The truth assignment function $m$ is a function from $\mathcal{P} \times W$ into the truth values $\{0, 1\}$. [3] The weight function $f$ is a function from $W$ to the set of real numbers $\Re$.*

The *valuation function* $V_P$ of the possible world partition sequence $P$ is constructed from $m$ in the usual way. Note that some of the $W_i$'s may be empty, but every possible world with a distinct truth assignment for the propositional constants has to be included in one of the classes. Also, the order of the classes in a partition sequence is significant. For example, $\langle \langle W_0, W_1 \rangle, m, f \rangle$ is distinct from $\langle \langle W_1, W_0 \rangle, m, f \rangle$, as the order of the classes in a partition sequence reflects the relative magnitude of bias of the worlds in the different classes.

---

[1]A partition of a set $S$ is a set of *non-empty* sets $S_1, \ldots, S_n$, such that $\bigcup_i S_i = S$, and $S_i \cap S_j = \emptyset$, for $i \neq j$.

The weight function $f$ denotes the *quantitative* bias on each world. In the cases when $f$ does not play a role in determining the results we are interested in, such as when the bias is induced by *qualitative* default rules, $f$ may be omitted. (We can assume that $f(w) = 1$ for all $w \in W$ if not specified.)

As a shorthand notation in examples, each world $w$ is denoted by an ordered pair $\langle S, f(w) \rangle$, such that $S$ is the set of propositional constants or their negations that are true in that world, and $f(w)$ is the weight assigned to the world. For example, $\langle \{p, \neg q\}, 0.2 \rangle$ represents a world in which $p$ is true but $q$ is false, and the weight is 0.2. Again, $f(w)$ may be omitted, and the world $\langle \{p, \neg q\}, 1 \rangle$ would be denoted by $\{p, \neg q\}$ in some cases. Similarly, a possible world partition sequence can be denoted simply by $\langle W_0, \ldots, W_l \rangle$, where each class $W_i$ is a set of possible world ordered pairs.

## 3 OVERVIEW OF SOME FORMALISMS

Let us briefly summarize the preliminary terminology and machinery of a few of the formalisms to uncertain reasoning. These include default logic, autoepistemic logic, probabilistic conditioning and thresholding, and possibility theory.

### 3.1 DEFAULT LOGIC [Reiter, 1980]

**Definition 3** *A default rule is an expression of the form $\frac{\alpha : \mathbf{M}\beta_1, \ldots, \mathbf{M}\beta_n}{\gamma}$, where $\alpha, \beta_1, \ldots, \beta_n$ and $\gamma$ are well formed formulas of $\mathcal{L}$. A default theory $\Delta$ is an ordered pair $\langle D, F \rangle$, where $D$ is a set of default rules and $F$ is a set of well formed formulas (facts) of $\mathcal{L}$.*

Intuitively, a default rule $\frac{\alpha : \mathbf{M}\beta_1, \ldots, \mathbf{M}\beta_n}{\gamma}$ represents that if $\alpha$ is provable, and $\neg \beta_1, \ldots, \neg \beta_n$ is each not provable, then we by default assert that $\gamma$ is true. For a default theory $\Delta = \langle D, F \rangle$, the known facts about the world constitute $F$, and a theory extended from $F$ by applying the default rules in $D$ is known as an *extension* of $\Delta$, defined as follows.

**Definition 4** *Let $\Delta = \langle D, F \rangle$ be a default theory over the language $\mathcal{L}$, and $E$ be a subset of $\mathcal{L}$. $\Gamma(E)$ is the smallest set satisfying the following three properties. [1] $F \subseteq \Gamma(E)$, [2] $\Gamma(E) = \mathbf{Th}(\Gamma(E))$, and [3] For every default rule $\frac{\alpha : \mathbf{M}\beta_1, \ldots, \mathbf{M}\beta_n}{\gamma} \in D$, if $\alpha \in \Gamma(E)$, and $\neg \beta_1, \ldots, \neg \beta_n \notin E$, then $\gamma \in \Gamma(E)$.*

$E$ is an extension of $\Delta$ iff $E$ is a fixed point of the operator $\Gamma$, that is, $E = \Gamma(E)$.

### 3.2 AUTOEPISTEMIC LOGIC

Given a set of premises $A \subseteq \mathcal{ML}$, an *autoepistemic theory* [Moore, 1985; Moore, 1984] $T \subseteq \mathcal{ML}$ is a set of modal formulas meant to be a set of beliefs of an agent when reflecting upon $A$. The principal modal operator of autoepistemic logic is $\mathbf{L}$, where $\mathbf{L}\phi$ is interpreted as that $\phi$ is believed. The belief set of an ideal agent is called a *stable expansion*, defined as follows.

**Definition 5** *Let $A \subseteq \mathcal{ML}$ be a set of premises and $T \subseteq \mathcal{ML}$ be an autoepistemic theory.*

- *$T$ is* stable *[Stalnaker, 1980] iff [1] if $\alpha_1, \ldots, \alpha_n \in T$, and $\alpha_1, \ldots, \alpha_n \vdash \beta$, then $\beta \in T$, [2] if $\alpha \in T$, then $\mathbf{L}\alpha \in T$, and [3] if $\alpha \notin T$, then $\neg \mathbf{L}\alpha \in T$.*

- *$T$ is* grounded *in $A$ iff every formula of $T$ is included in the tautological consequences of $A \cup \{\mathbf{L}\alpha : \alpha \in T\} \cup \{\neg\mathbf{L}\alpha : \alpha \notin T\}$.*

*$T$ is a* stable expansion *of $A$ iff [1] $A \subseteq T$, [2] $T$ is stable, and [3] $T$ is grounded in $A$.*

One useful property of stability is that each stable autoepistemic theory is uniquely determined by its *kernel*, the set of non-modal formulas in the theory. Thus, we only need to specify the kernel when we refer to stable theories.

Konolige [Konolige, 1988; Konolige, 1989] showed that in the modal system **K45**, every well formed formula of $\mathcal{ML}$ is equivalent to a formula of the *normal form* $\neg\mathbf{L}\alpha \vee \mathbf{L}\beta_1 \vee \ldots \vee \mathbf{L}\beta_n \vee \gamma$, where $\alpha, \beta_1, \ldots, \beta_n, \gamma$ are all non-modal formulas, and any of $\alpha, \beta_1, \ldots, \beta_n$ may be absent[2]. We assume that all autoepistemic formulas are given in an equivalent normal form $\mathbf{L}\alpha \wedge \neg\mathbf{L}\beta_1 \wedge \ldots \wedge \neg\mathbf{L}\beta_n \rightarrow \gamma$ in our discussion.

### 3.3 PROBABILISTIC CONDITIONING AND THRESHOLDING

A probability function Pr is characterized by the following three axioms[3]. For any events $E_1$ and $E_2$ in the field of $S$, [1] $0 \leq \Pr(E_1) \leq 1$, [2] $\Pr(S) = 1$, and [3] $\Pr(E_1 \cup E_2) = \Pr(E_1) + \Pr(E_2)$ if $E_1 \cap E_2 = \emptyset$.

We focus on two types of probability operations, *conditioning* and *thresholding*. $\Pr(\psi \mid \phi)$, the probability of $\psi$ conditioned on $\phi$, is given by $\frac{\Pr(\psi \wedge \phi)}{\Pr(\phi)}$ when $\Pr(\phi) \neq 0$. This quantity can be computed for any pair of formulas $\phi$ and $\psi$ as long as the probability of $\phi$ is positive. However, the conditional probabilities of interest are typically those computed in response to a change of $\Pr(\phi)$ to 1. The probabilities of all formulas are then updated to take into account this change.

Thresholding is the process of accepting statements whose probability values exceed a certain threshold. This can be taken as a form of generalized conditioning. In regular conditioning, the conditional probability is computed when $\Pr(\phi)$ changes to 1. Thresholding imposes a weaker requirement[4]: accept $\phi$ if $\Pr(\phi) \geq 1 - \epsilon$. It provides an explanation of how we come to ignore certain improbable events, such as my ducks suffocating due to all air molecules simultaneously rushing to the far end of the cage. The probability of this scenario is smaller than any reasonable $\epsilon$, and so we take it as practically false[5].

**Definition 6** *The probability of $\psi$ thresholded at $\epsilon$ wrt $\langle \phi_1, \ldots, \phi_n \rangle$, $n \geq 1$, denoted by $\Pr_\epsilon(\psi \mid \langle \phi_1, \ldots, \phi_n \rangle)$, is $\Pr(\psi \mid \phi_1, \ldots, \phi_n)$ iff for $1 \leq i \leq n$, $\Pr(\phi_i \mid \phi_1, \ldots, \phi_{i-1}) \geq 1 - \epsilon$.*

Note that the effective probability space is shrinking. After $\phi_1$ is thresholded, we only consider the space in which $\phi_1$ is true in future computations, and the revised probability of all formulas $\psi$ becomes $\Pr(\psi \mid \phi_1)$, which becomes $\Pr(\psi \mid \phi_1, \phi_2, \ldots)$ as more formulas are thresholded. Since in general $\Pr(\psi \mid \phi_1, \ldots, \phi_k) \neq \Pr(\psi \mid \phi_1, \ldots, \phi_k, \phi_{k+1})$, each of the $\phi_i$'s to be thresholded is required to have a probability that is above threshold at the time it is treated. Also note that a set of formulas may be thresholded in multiple ways, depending on the specific formula picked at each stage.

### 3.4 POSSIBILITY THEORY

**Definition 7** *The* possibility measure $\Pi$ *and* necessity measure $N$ *satisfy the following axioms [Zadeh, 1978; Dubois and Prade, 1988]. For any formulas $\phi$ and $\psi$, [1] $\Pi(\bot) = N(\bot) = 0$, [2] $\Pi(\top) = N(\top) = 1$, [3] $\Pi(\phi \vee \psi) = \max(\Pi(\phi), \Pi(\psi))$, and [4] $N(\phi \wedge \psi) = \min(N(\phi), N(\psi))$. The standard measures in addition bears the relation $N(\phi) = 1 - \Pi(\neg\phi)$.*

We consider only a simplified version of possibility here. The underlying concepts are crisp (each proposition is either totally true or totally false), but we do not have sufficient information to determine with complete confidence one way or the other. Also the knowledge base is coherent, in the sense that for any formula $\phi$, at least one of the necessity values $N(\phi)$ and $N(\neg\phi)$ is 0. We cannot have assertions that simultaneously support the necessity of a set of outcomes and its negation[6].

---

[2] Note that the non-modal disjunct $\gamma$ has to be present, and thus the normal form of the formula $\mathbf{L}\alpha$ is $\mathbf{L}\alpha \vee \bot$.

[3] It suffices to specify the third property as finite additivity instead of the more general countable additivity, since we are considering only a finite space.

[4] The $\epsilon$ in the threshold does not tend to 0, as in $\epsilon$-semantics [Adams, 1975; Pearl, 1989], but is assumed to be fixed, although it can vary with context.

[5] Note that a statement $\phi$ is taken to be false if $\Pr(\phi) \leq \epsilon$, not if $\Pr(\phi) < 1 - \epsilon$.

[6] This is the basic scenario. Possibility theory stems

## 4  SPECIFIC PARTITION SEQUENCES

Now we show how the formalisms described in the previous section can be characterized using the possible world partition sequence framework.

### 4.1  DEFAULT PARTITION SEQUENCE

**Definition 8** *A possible world partition sequence $P = \langle\langle W_0,\ldots,W_l\rangle, m\rangle$ is a default partition sequence for a default theory $\Delta = \langle D, F\rangle$ iff it satisfies the following properties.*

1. $W_0 = \{w \in \bigcup_i W_i : V_P(F, w) = 0\}$[7].

2. *For each $W_i$, $0 < i < l$, there exists a default rule $\frac{\alpha : \mathbf{M}\beta_1,\ldots,\mathbf{M}\beta_n}{\gamma} \in D$, such that [1] $V_P(\alpha, w) = 1$ for all $w \in W_i,\ldots,W_l$, [2] $\exists w_1,\ldots,w_n \in W_l$ such that $V_P(\beta_1, w_1) = 1,\ldots,V_P(\beta_n, w_n) = 1$, and [3] $W_i = \{w \notin W_0,\ldots,W_{i-1} : V_P(\gamma, w) = 0\}$.*

3. *For all default rules $\frac{\alpha : \mathbf{M}\beta_1,\ldots,\mathbf{M}\beta_n}{\gamma} \in D$, if $V_P(\bullet, w) = 1$ for all $w \in W_l$, and $\exists w_1,\ldots,w_n \in W_l$ such that $V_P(\beta_1, w_1) = 1,\ldots,V_P(\beta_n, w_n) = 1$, then $V_P(\gamma, w) = 1$ for all $w \in W_l$.*

**Theorem 9** *A set of formulas $E$ is an extension of a default theory $\Delta = \langle D, F\rangle$ iff there is a default partition sequence $P = \langle\langle W_0,\ldots,W_l\rangle, m\rangle$ for $\Delta$, such that $E$ is the set of formulas $\{\phi : V_P(\phi, w) = 1, \forall w \in W_l\}$.*

A default partition sequence captures the successive selection of possible worlds by the default rules. The first class $W_0$ consists of all those worlds that are not consistent with the given facts $F$. At each level $0 < i < l$, an applicable default rule is chosen, and all the worlds that are not consistent with its default conclusion $\gamma$ are grouped into class $W_i$, which are ignored when further default rules are applied. Thus, the worlds in the tail of the sequence, $W_{i+1}\ldots W_l$, are the worlds that are still "suitable by default" after $i$ default rules have been applied. No more default rule is applicable to the final class $W_l$, and so it consists of all the possible worlds of the extension.

The mapping from default partition sequences to default extensions is many-to-one. There can be more than one partition sequence for each extension, depending on, for instance, the order in which default rules are applied when multiple rules are applicable at some stage. Two different default partition sequences are isomorphic iff their last class $W_l$ are identical.

---

from fuzzy set theory, and can be applied to fuzzy sets as well as to crisp sets. See also [Dubois *et al.*, 1994] for the treatment of partially inconsistent knowledge bases.

[7]We assume here that the set of formulas $F = \{f_1, f_2, \ldots\}$ is equivalent to the single formula $F = f_1 \wedge f_2 \wedge \ldots$.

The following examples illustrate the correspondence between extensions of a default theory and their default partition sequences.

**Example 10** $D = \{\frac{:\mathbf{M}p}{p}, \frac{:\mathbf{M}\neg p}{\neg p}, \frac{p:\mathbf{M}q}{q}\}$, and $F = \emptyset$.

There are two extensions, $E_1 = \mathbf{Th}(\{\neg p\})$ and $E_2 = \mathbf{Th}(\{p, q\})$, which can be characterized by the two default partition sequences [1] $\langle \emptyset, W_{11}, W_{12}\rangle$, where $W_{11} = \{\{p, q\}, \{p, \neg q\}\}$, $W_{12} = \{\{\neg p, q\}, \{\neg p, \neg q\}\}$, and [2] $\langle \emptyset, W_{21}, W_{22}, W_{23}\rangle$, where $W_{21} = \{\{\neg p, q\}, \{\neg p, \neg q\}\}$, $W_{22} = \{\{p, \neg q\}\}$, $W_{23} = \{\{p, q\}\}$. $\square$

**Example 11** $\Delta = \langle D, \emptyset\rangle$, where $D = \{\frac{:\mathbf{M}p}{\neg p}\}$.

This default theory has no extension. To construct any default partition sequence for $\Delta$, $W_0$ has to be $\emptyset$, since $F$ is empty. There are two possible cases. [1] $\{p\} \in W_l$. But then $\neg p$ has to be true in all worlds in $W_l$ according to condition 3 in Definition 8. [2] $\{p\} \notin W_l$. Then the default rule is not applicable, and we cannot construct intermediate classes for $\{p\}$ according to condition 2. $\square$

### 4.2  AUTOEPISTEMIC PARTITION SEQUENCE

**Definition 12** *A possible world partition sequence $P = \langle\langle W_0,\ldots,W_l\rangle, m\rangle$ is an autoepistemic partition sequence for $A \subseteq \mathcal{ML}$ iff it satisfies the following properties.*

1. $W_0 = \emptyset$.

2. *For each $W_i$, $0 < i < l$, there exists a formula $\mathbf{L}\alpha \wedge \neg\mathbf{L}\beta_1 \wedge \ldots \wedge \neg\mathbf{L}\beta_n \rightarrow \gamma \in A$ such that [1] $V_P(\alpha, w) = 1$ for all $w \in W_l$, [2] $\exists w_1,\ldots,w_n \in W_l$ such that $V_P(\neg\beta_1, w_1) = 1,\ldots,V_P(\neg\beta_n, w_n) = 1$ and [3] $W_i = \{w \notin W_0,\ldots,W_{i-1} : V_P(\gamma, w) = 0\}$.*

3. *For all formulas $\mathbf{L}\alpha \wedge \neg\mathbf{L}\beta_1 \wedge \ldots \wedge \neg\mathbf{L}\beta_n \rightarrow \gamma \in A$, if $V_P(\alpha, w) = 1$ for all $w \in W_l$, and $\exists w_1,\ldots,w_n \in W_l$ such that $V_P(\neg\beta_1, w_1) = 1,\ldots,V_P(\neg\beta_n, w_n) = 1$, then $V_P(\gamma, w) = 1$ for all $w \in W_l$.*

**Theorem 13** *An autoepistemic theory $T$ is a consistent stable expansion of a set of premises $A$ iff there is an autoepistemic partition sequence $P = \langle\langle W_0,\ldots,W_l\rangle, m\rangle$ for $A$, such that $W_l \neq \emptyset$ and $T$ is the stable set characterized by the kernel $\{\phi : V_P(\phi, w) = 1, \forall w \in W_l\}$.*

The definition of an autoepistemic partition sequence is very similar to that of a default partition sequence, with differences parallel to those occurring in their fixed point formulations [Teng, 1996][8]. In particular,

---

[8]The first condition in Definition 12 is by no means essential. It is added so that the parallel between default

in item 2 of Definition 12, $\alpha$ has to be true only in the worlds in the last class $W_l$, while in the corresponding condition in Definition 8, $\alpha$ is evaluated against all classes $W_i, \ldots, W_l$.

**Example 14** $A = \{\mathbf{L}p \to p, \neg \mathbf{L}p \to q\}$.

There are two stable expansions, $T_1$ with the kernel $\{p\}$ and $T_2$ with the kernel $\{q\}$. An autoepistemic partition sequence corresponding to $T_1$ is $\langle \emptyset, W_{11}, W_{12}\rangle$, with $W_{11} = \{\{\neg p, q\}, \{\neg p, \neg q\}\}$, $W_{12} = \{\{p, q\}, \{p, \neg q\}\}$, and $\mathbf{L}p \to p$ is used for the partition. A partition sequence corresponding to $T_2$ is $\langle \emptyset, W_{21}, W_{22}\rangle$, with $W_{21} = \{\{p, \neg q\}, \{\neg p, \neg q\}\}$, $W_{22} = \{\{p, q\}, \{\neg p, q\}\}$, and the formula $\neg \mathbf{L}p \to q$ is used for this partition. $\square$

**Example 15** $A = \{\neg \mathbf{L}p \to q, \neg q\}$.

$A$ has no stable expansion. To construct any autoepistemic partition sequence, $W_l$ can only contain one or both of the worlds $\{p, \neg q\}, \{\neg p, \neg q\}$, since $\neg q \in A$ and by condition 3 of Definition 12, $\neg q$ is true in all worlds in $W_l$. If $W_l = \{\{p, \neg q\}\}$, there is no formula that can be used to construct a class for $\{\neg p, \neg q\}$ according to condition 2. If $W_l = \{\{\neg p, \neg q\}\}$ or $\{\{p, \neg q\}, \{\neg p, \neg q\}\}$, then $q$ needs to be true in all worlds in $W_l$ by condition 3. $\square$

### 4.3 CONDITIONAL AND THRESHOLD PARTITION SEQUENCES

The sample space can be represented as a set of possible worlds $\Delta = \langle W, m, f\rangle$. For example, a sample space for fair coin tosses is $\{\langle\{h\}, 0.5\rangle, \langle\{\neg h\}, 0.5\rangle\}$, where the proposition $h$ stands for heads.

**Definition 16** *A possible world partition sequence $P = \langle\langle W_0, \ldots, W_n\rangle, m, f\rangle$ is a conditional probability partition sequence for a sample space $\Delta = \langle W, m, f\rangle$ conditioned on a sequence of formulas $\langle\phi_1, \ldots, \phi_n\rangle$, $n \geq 1$, iff [1] $\bigcup_{0 \leq i \leq n} W_i = W$, and [2] $W_i = \{w \notin W_0, \ldots, W_{i-1} : V_P(\phi_{i+1}, w) = 0\}$ for $0 \leq i < n$.*

**Theorem 17** *Given a sample space $\Delta = \langle W, m, f\rangle$, $\Pr(\psi \mid \phi_1, \ldots, \phi_n) = r$ iff there is a conditional probability partition sequence $P = \langle\langle W_0, \ldots, W_n\rangle, m, f\rangle$ for $\Delta$ conditioned on $\langle\phi_1, \ldots, \phi_n\rangle$, and $\frac{\sum_{w \in W_n : V_P(\psi, w) = 1} f(w)}{\sum_{w \in W_n} f(w)} = r$, where $\sum_{w \in W_n} f(w) \neq 0$.*

**Example 18** $\Delta = \{w_1, w_2, w_3, w_4\}$, where $w_1 = \langle\{p, q\}, 0.2\rangle$, $w_2 = \langle\{p, \neg q\}, 0.3\rangle$, $w_3 = \langle\{\neg p, q\}, 0.1\rangle$, $w_4 = \langle\{\neg p, \neg q\}, 0.4\rangle$.

A conditional probability partition sequence for $\Delta$ conditioned on $\langle p \to q, p \vee q\rangle$ is $\langle\{w_2\}, \{w_4\}, \{w_1, w_3\}\rangle$. Then we have $\Pr(p \mid p \to q, p \vee q) = \frac{f(w_1)}{f(w_1) + f(w_3)} = \frac{2}{3}$, and $\Pr(q \mid p \to q, p \vee q) = \frac{f(w_1) + f(w_3)}{f(w_1) + f(w_3)} = 1$. $\square$

---

and autoepistemic partition sequences can be brought out more clearly.

The conditioning probability partition sequence is incremental in the sense that we can build a partition sequence conditioned on $\langle\phi_1, \ldots, \phi_{k+1}\rangle$ based on a partition sequence conditioned on $\langle\phi_1, \ldots, \phi_k\rangle$. It is also persistent [Driscoll *et al.*, 1989] since we can retrieve $\Pr(\psi \mid \phi_1, \ldots, \phi_k)$ from a partition sequence conditioned on $\langle\phi_1, \ldots, \phi_k, \phi_{k+1}, \ldots\rangle$.

Now we turn to thresholding.

**Definition 19** *A possible world partition sequence $P = \langle\langle W_0, \ldots, W_n\rangle, m, f\rangle$ is a threshold probability partition sequence for a sample space $\Delta = \langle W, m, f\rangle$ at $\epsilon$ wrt $\langle\phi_1, \ldots, \phi_n\rangle$ iff [1] $P$ is a conditional probability partition sequence for $\Delta$ conditioned on $\langle\phi_1, \ldots, \phi_n\rangle$, and [2] $\frac{\sum_{w \in W_i} f(w)}{\sum_{w \in W_i, \ldots, W_n} f(w)} \leq \epsilon$ for $0 \leq i < n$.*

Condition 2 gives an equivalent condition for $\phi_{i+1}$ to be above threshold: the weighted proportion of worlds in $W_i$ (those worlds in which $\phi_{i+1}$ is false) among all worlds in $W_i, \ldots, W_n$ is no greater than $\epsilon$.

**Theorem 20** *Given a sample space $\Delta = \langle W, m, f\rangle$, $\Pr_\epsilon(\psi \mid \langle\phi_1, \ldots, \phi_n\rangle) = r$ iff there is a threshold probability partition sequence $P = \langle\langle W_0, \ldots, W_n\rangle, m, f\rangle$ for $\Delta$ at $\epsilon$ wrt $\langle\phi_1, \ldots, \phi_n\rangle$, and $\frac{\sum_{w \in W_n : V_P(\psi, w) = 1} f(w)}{\sum_{w \in W_n} f(w)} = r$, where $\sum_{w \in W_n} f(w) \neq 0$.*

#### 4.3.1 The Lottery Paradox: An Example

The lottery paradox [Kyburg, 1961] is as follows. 100 tickets have been sold, but only one will win. Each ticket has an equal but small chance, and thus can be regarded as practically losing. However, if we apply this train of thought to all tickets, we end up rejecting all 100 tickets as losing, which is inconsistent with the premise that one of the tickets will win. We can formulate the lottery paradox using thresholding.

**Example 21** Let $\Delta = \langle W, m, f\rangle$, where $w_i = \langle\{\neg p_1, \ldots, \neg p_{i-1}, p_i, \neg p_{i+1}, \ldots, \neg p_{100}\}, \frac{1}{100}\rangle$, and $W = \{w_1, \ldots, w_{100}\}$[9]. $P = \langle W_0, W_1, W_2\rangle$, where $W_0 = \{w_1\}, W_1 = \{w_2\}, W_2 = \{w_3, \ldots, w_{100}\}$, is a threshold probability partition sequence for $\Delta$ at $\epsilon = \frac{1}{99}$ wrt $\langle\neg p_1, \neg p_2\rangle$. We can check that $P$ satisfies the conditions in Definition 19. [1] $P$ is a conditional probability partition sequence conditioned on $\langle\neg p_1, \neg p_2\rangle$. [2] $\frac{\sum_{w \in W_0} f(w)}{\sum_{w \in W_0, W_1, W_2} f(w)} = \frac{f(w_1)}{f(w_1) + \ldots + f(w_{100})} = \frac{1}{100}$. Similarly, $\frac{\sum_{w \in W_1} f(w)}{\sum_{w \in W_1, W_2} f(w)} = \frac{f(w_2)}{f(w_2) + \ldots + f(w_{100})} = \frac{1}{99}$. Both of these proportions are $\leq \epsilon$.

Thus, we can derive $\Pr(p_1) = \Pr(p_2) = 0$, and $\Pr(p_i) = \frac{1}{98}$ for all $i \geq 3$, since $p_1$ and $p_2$ are both false in all the worlds in $W_2$, and for each $p_i$, $i \geq 3$, there is exactly one world among the 98 in the last class $W_2$ at which $p_i$ is true. □

---

[9]All other worlds have a weight of 0, and thus can be safely ignored since they do not contribute to the weight of any set.

Each proposition $p_i$ corresponds to the statement "ticket $i$ wins", and each world $w_i$ corresponds to the situation in which ticket $i$ wins and all others lose. In the partition sequence constructed in the example, the first and second tickets are deemed losing, but nothing certain is concluded about the remaining tickets. In fact, for any two arbitrary tickets $i$ and $j$, we can construct a threshold probability partition sequence for $\Delta$ at $\frac{1}{99}$ wrt $\langle \neg p_i, \neg p_j \rangle$, and tickets $i$ and $j$ are losing in this situation.

Note that the last class $W_2$ cannot be further partitioned at $\epsilon = \frac{1}{99}$ by thresholding on additional formulas, since if we were to split $W_2$ into two non-empty classes $W_2'$ and $W_3'$, $\frac{\sum_{w \in W_2'} f(w)}{\sum_{w \in W_2' \cdot W_3'} f(w)}$ is at least $\frac{0.01}{0.98}$ (as there are 98 worlds in $W_2$), which is greater than $\epsilon$. Thus, the threshold $1 - \epsilon$ represents the limit to what we consider as "practically true". With $\epsilon = \frac{1}{99}$, we are willing to say that a person who holds 2, but not 3, tickets will lose, while with $\epsilon = \frac{1}{100}$, we only commit to saying any one ticket will lose. The uncomfortable conclusion that all tickets will lose occurs when $\epsilon = 1$. This allows us to threshold, and take as true, any formula with probability $\geq 0$, which amounts to all formulas in $\mathcal{L}$, and thus the inconsistency. To avoid the lottery paradox, we only need to set $\epsilon < 1$.

### 4.4 POSSIBILITY PARTITION SEQUENCE

Consider a set of possibilistic statements $S = \{\langle S_1, r_1 \rangle, \ldots, \langle S_n, r_n \rangle\}$, $n \geq 1$, where $S_i \neq \emptyset$ is a set of well formed formulas of $\mathcal{L}$, and $\Pi(\phi) = r_i$ for all $\phi \in S_i$. Without loss of generality, we assume $r_1 < \ldots < r_n$.

**Definition 22** *A possible world partition sequence $P = \langle \langle W_0, \ldots, W_n \rangle, m, f \rangle$ is a possibility partition sequence for a set of consistent possibilistic statements $S = \{\langle S_1, r_1 \rangle, \ldots, \langle S_n, r_n \rangle\}$ iff [1] For $0 \leq i < n$, $W_i = \bigcup_{\phi \in S_{i+1}} U_\phi$, where $U_\phi = \{w \notin W_0, \ldots, W_{i-1} : V_P(\phi, w) = 1\}$ and all $U_\phi \neq \emptyset$ unless $r_i = 0$, and [2] $\sum_{w \in W_i} f(w) = r_{i+1} - r_i$, assuming $r_0 = 0$ and $r_{n+1} = 1$.*

Note that the $U_\phi$'s need not be disjoint (but all must be non-empty). Only the "meta"-classes $W_i$'s formed from their unions need to be disjoint. Also the weights of individual worlds can vary as long as the total weight in each class satisfies the possibilistic constraints.

**Theorem 23** *Given a consistent set of possibilistic statements $S = \{\langle S_1, r_1 \rangle, \ldots, \langle S_n, r_n \rangle\}$, $\Pi(\phi) = r$ iff there is a possibility partition sequence $P = \langle \langle W_0, \ldots, W_n \rangle, m, f \rangle$ for $S$, and $\sum_{w \in W_0, \ldots, W_i} f(w) = r$, where $i = \max(\{k : \exists w \in W_k, V_P(\phi, w) = 1\})$. If there is no world $w$ such that $V_P(\phi, w) = 1$, then $\Pi(\phi) = 0$.*

In a possibility partition sequence, the possible worlds are divided into classes $W_0, \ldots, W_n$ of increasing possibility. Formulas that are true in any of the worlds in $W_i$, but not true in any world in $W_{i+1}, \ldots, W_n$, are possible to the same extent. The weights of all the worlds add up to 1, and so the possibility of formulas that are true in any world in the final class $W_n$ is always 1. The order of the classes in the partition sequence represents the successive grouping of worlds in accordance to how possible they are. $W_0$ contains the worlds that are least possible, and $W_n$ contains those worlds that are most possible.

**Example 24** $S = \{\langle \{p \wedge q\}, 0.3 \rangle, \langle \{p\}, 0.7 \rangle\}$.

$P = \langle \langle W_0, W_1, W_2 \rangle, m, f \rangle$ is a possibility partition sequence for $S$, where $W_0 = \{\langle \{p, q\}, 0.3 \rangle\}$, $W_1 = \{\langle \{p, \neg q\}, 0.4 \rangle\}$, $W_2 = \{\langle \{\neg p, q\}, k \rangle, \langle \{\neg p, \neg q\}, 0.3 - k \rangle\}$ with $0 \leq k \leq 0.3$.

We can verify that $\Pi(p) = 0.7$, as stated in $S$. $p$ is true in a world in $W_1$ but not true in any of the worlds in $W_2$, and therefore $\Pi(p)$ is given by $0.3 + 0.4 = 0.7$. $\Pi(q) = \Pi(\neg q) = 1$, since $q$ and $\neg q$ each is true in some (different) world in $W_2$. This amounts to saying that we have no information regarding the truth of $q$. □

**Example 25** $S = \{\langle \{p\}, 0.3 \rangle, \langle \{p \wedge q\}, 0.5 \rangle\}$.

$S$ is inconsistent, since on one hand $\Pi(p) = 0.3$ as stated in $S$, but we can also derive $\Pi(p)$ by noting that $\Pi(p) = \Pi((p \wedge q) \vee (p \wedge \neg q)) \geq \Pi(p \wedge q) = 0.5$. To construct a possibility partition sequence, we would have $P = \langle \langle W_0, W_1, W_2 \rangle, m, f \rangle$ for $S$, where $W_0 = \{\langle \{p, q\}, k_0 \rangle, \langle \{p, \neg q\}, 0.3 - k_0 \rangle\}$, $W_1 = \emptyset$, $W_2 = \{\langle \{\neg p, q\}, k_2 \rangle, \langle \{\neg p, \neg q\}, 0.5 - k_2 \rangle\}$, with $0 \leq k_0 \leq 0.3$ and $0 \leq k_2 \leq 0.5$. The set $U_{p \wedge q}$ for constructing $W_1$ is empty, which does not satisfy condition 1 of Definition 22. Note also that the weights of all the worlds in $P$ do not add up to 1. □

## 5 CONCLUSION

We presented a framework for unifying formalisms of uncertain reasoning. The framework is based on an ordered partition of possible worlds we call partition sequences. A partition sequence corresponds to our intuitive notion of biasing towards certain possible scenarios when we are uncertain of the actual situation. The constraints for constructing allowable partition sequences reflect the characterizing way the bias is assigned in a formalism. We showed that default logic, autoepistemic logic, probabilistic conditioning and thresholding, and possibility theory can be successfully assimilated into this framework. As a side point, we also showed how the lottery paradox can be avoided by probabilistic thresholding, and how it

Possible World Partition Sequences 523

can be expressed as a threshold probability partition sequence.

The semantics we provide is similar in flavor to Shoham's preference semantics [Shoham, 1988]. Instead of imposing a preference ordering on models, we impose an ordering on equivalence classes of possible worlds. A possible world partition sequence $\langle W_0, \ldots, W_n \rangle$ can be regarded in a broad sense as a preference relation of the models $M_i = \bigcup_{i \leq k \leq n} W_k$ so that $M_0 \sqsubset \ldots \sqsubset M_n$, where $M_i \sqsubset M_j$ is interpreted as that $M_j$ is preferred over $M_i$.

Our work presented here provides a common framework in which we can characterize various formalisms to uncertain reasoning. We consider it the ground work for building an integrated system with a well founded semantics on the mechanism of combining knowledge bases of multiple formats.

## Acknowledgements

I would like to thank my advisor Henry Kyburg for his inexhaustible patience and good advice. This work was supported by National Science Foundation grant IRI-9411267.

## APPENDIX: PROOFS OF THEOREMS

We append here sketches of proofs of the theorems presented in this paper.

**Theorem 9** *A set of formulas $E$ is an extension of a default theory $\Delta = \langle D, F \rangle$ iff there is a default partition sequence $P = \langle \langle W_0, \ldots, W_l \rangle, m \rangle$ for $\Delta$, such that $E$ is the set of formulas $\{\phi : V_P(\phi, w) = 1, \forall w \in W_l\}$.*

**Proof** ($\Longrightarrow$) Suppose $E$ is an extension of a default theory $\Delta = \langle D, F \rangle$, that is, $E = \Gamma(E)$. We need to show that there is a default partition sequence $P = \langle \langle W_0, \ldots, W_l \rangle, m \rangle$ for $\Delta$, such that $E = \{\phi : V_P(\phi, w) = 1, \forall w \in W_l\}$.

Let $W_l$ be the set of all possible worlds in which $E$ is true. Condition 3 in Definition 8 is satisfied as a consequence of condition 3 of Definition 4. In addition, we can always construct $W_0$ according to condition 1, and order the remaining worlds into the sequence $\langle W_1, \ldots, W_m \rangle$ according to condition 2 of Definition 8, until $W_m$ cannot be further partitioned. Now we only need to show that $P = \langle\langle W_0, \ldots, W_{m-1}, W_l\rangle, m\rangle$ is a partition sequence, that is, $W_m = W_l$. We proceed in two steps.

The first two items in condition 2 of Definition 8 entails the antecedent in condition 3 of Definition 4, and so $\gamma \in E$ and $V_P(\gamma, w) = 1$ for all $w \in W_l$. Therefore none of the worlds in $W_l$ can be grouped into any of the $W_i$'s, $i < m$, and $W_l \subseteq W_m$.

Now we show that it is not the case that $W_l \subset W_m$. Assume the contrary. Let $E' = \{\phi : V_P(\phi, w) = 1, \forall w \in W_m\}$. Recall that $W_l$ is a maximal set for $E$, and the additional worlds in $W_m$ makes $E' \subset E$. $E'$ also satisfies the three conditions in Definition 4 as a candidate for $\Gamma(E)$, which contradicts that $E$ is an extension and thus the *smallest* such candidate.

($\Longleftarrow$) Suppose there is a default partition sequence $P = \langle\langle W_0, \ldots, W_l\rangle, m\rangle$ for $\Delta = \langle D, F\rangle$. We need to show that $E = \{\phi : V_P(\phi, w) = 1, \forall w \in W_l\}$ is an extension of $\Delta$, that is, $E = \Gamma(E)$.

$E$ satisfies the three conditions in Definition 4, and thus $\Gamma(E) \subseteq E$. Now let $E_i = \{\phi : V_P(\phi, w) = 1, \forall w \in W_i, \ldots, W_l\}$. We show $E_i \subseteq \Gamma(E)$ for $i > 0$. $E_1 = F \subseteq \Gamma(E)$. Now assume $E_i \subseteq \Gamma(E)$. $E_{i+1} = \mathbf{Th}(E_i \cup \{\gamma\})$, where there is a default rule $\frac{\alpha : \mathbf{M}\beta_1, \ldots, \mathbf{M}\beta_n}{\gamma} \in D$, such that [1] $V_P(\alpha, w) = 1$ for all $w \in W_i, \ldots, W_l$, and [2] $\exists w_1, \ldots, w_n \in W_l$ such that $V_P(\beta_1, w_1) = 1, \ldots, V_P(\beta_n, w_n) = 1$. From [1] we have $\alpha \in \Gamma(E)$, and from [2] $\neg\beta_1, \ldots, \neg\beta_n \notin E$. Thus, $\gamma \in \Gamma(E)$ according to condition 3 of Definition 4, and $E_{i+1} \subseteq \Gamma(E)$. In particular, $E = E_l \subseteq \Gamma(E)$. □

**Theorem 13** *An autoepistemic theory $T$ is a consistent stable expansion of a set of premises $A$ iff there is an autoepistemic partition sequence $P = \langle\langle W_0, \ldots, W_l\rangle, m\rangle$ for $A$, such that $W_l \neq \emptyset$ and $T$ is the stable set characterized by the kernel $\{\phi : V_P(\phi, w) = 1, \forall w \in W_l\}$.*

**Proof** We make use of the following Lemma.

**Lemma 13.1** *[Teng, 1996] Let $A \subseteq \mathcal{ML}$ be a set of formulas (premises) in normal form and $T$ be a consistent subset of $\mathcal{ML}$. $\Omega(T)$ is the set with the smallest kernel satisfying the following two properties. [1] $\Omega(T)$ is stable, and [2] For every formula $\mathbf{L}\alpha \wedge \neg\mathbf{L}\beta_1 \wedge \ldots \wedge \neg\mathbf{L}\beta_n \to \gamma \in A$, if $\alpha \in T$, and $\beta_1, \ldots, \beta_n \notin T$, then $\gamma \in \Omega(T)$. $T$ is a consistent stable expansion of $A$ iff $T$ is a fixed point of the operator $\Omega$, that is, $T = \Omega(T)$.*

The proof for autoepistemic partition sequences follows closely the one for default partition sequences (Theorem 9), with obvious adjustments corresponding to the differences between the fixed point formulations of the two logics. □

**Theorem 17** *Given a sample space $\Delta = \langle W, m, f\rangle$, $\Pr(\psi \mid \phi_1, \ldots, \phi_n) = r$ iff there is a conditional probability partition sequence $P = \langle\langle W_0, \ldots, W_n\rangle, m, f\rangle$ for $\Delta$ conditioned on $\langle\phi_1, \ldots, \phi_n\rangle$, and $\frac{\sum_{w \in W_n : V_P(\psi, w) = 1} f(w)}{\sum_{w \in W_n} f(w)} = r$, where $\sum_{w \in W_n} f(w) \neq 0$.*

**Proof** This theorem follows from the definition of conditional probabilities. □

**Theorem 20** *Given a sample space $\Delta = \langle W, m, f\rangle$, $\Pr_\epsilon(\psi \mid \langle\phi_1, \ldots, \phi_n\rangle) = r$ iff there is a threshold probability partition sequence $P = \langle\langle W_0, \ldots, W_n\rangle, m, f\rangle$ for $\Delta$ at $\epsilon$ wrt $\langle\phi_1, \ldots, \phi_n\rangle$, and $\frac{\sum_{w \in W_n : V_P(\psi, w) = 1} f(w)}{\sum_{w \in W_n} f(w)} = r$, where $\sum_{w \in W_n} f(w) \neq 0$.*

**Proof** This theorem follows from Theorem 17. Condition 2 in Definition 19 corresponds to the constraint $\Pr(\phi_i \mid \phi_1, \ldots, \phi_{i-1}) \geq 1 - \epsilon$ for $1 \leq i \leq n$ in Definition 6. □

**Theorem 23** *Given a consistent set of possibilistic statements $S = \{\langle S_1, r_1\rangle, \ldots, \langle S_n, r_n\rangle\}$, $\Pi(\phi) = r$ iff there is a possibility partition sequence $P = \langle\langle W_0, \ldots, W_n\rangle, m, f\rangle$ for $S$, and $\sum_{w \in W_0, \ldots, W_i} f(w) = r$, where $i = \max(\{k : \exists w \in W_k, V_P(\phi, w) = 1\})$. If there is no world $w$ such that $V_P(\phi, w) = 1$, then $\Pi(\phi) = 0$.*

**Proof** ($\Longrightarrow$) Given a consistent set of possibilistic statements $S = \{\langle S_1, r_1\rangle, \ldots, \langle S_n, r_n\rangle\}$, we can construct a possibility partition sequence $P = \langle\langle W_0, \ldots, W_n\rangle, m, f\rangle$ for $S$. We show that $U_\phi \neq \emptyset$ unless $r_i = 0$ in condition 1 of Definition 22. If $\phi$ is inconsistent, then $r_i$ must be 0 by Definition 7. Now suppose $\phi \in S_{i+1}$ is consistent but $U_\phi = \emptyset$. That is, the worlds in which $\phi$ is true are all in $W_0, \ldots, W_{i-1}$. In other words, there is a formula $\psi = \psi_1 \vee \ldots \vee \psi_k$ such that each $\psi_j \in S_0 \cup \ldots \cup S_i$, and $\psi \vdash \phi$. We have $\Pi(\psi) = \max(\Pi(\psi_1), \ldots, \Pi(\psi_k))$, which is $\leq r_i$. However, $\psi$ can be rewritten as $\psi_1 \vee \ldots \vee \psi_k \vee \phi$, and we have $\Pi(\psi) = \max(\Pi(\psi_1), \ldots, \Pi(\psi_k), \Pi(\phi))$, which is $\geq \Pi(\phi) = r_{i+1} > r_i$, a contradiction.

($\Longleftarrow$) Suppose there is a possibility partition sequence $P = \langle\langle W_0, \ldots, W_n\rangle, m, f\rangle$ for a set of possibilistic statements $S = \{\langle S_1, r_1\rangle, \ldots, \langle S_n, r_n\rangle\}$. Clearly the possibility values derived by the theorem satisfies the requirements in Definition 7. In addition, the statements in $S$ have their intended possibility values. For a formula $\phi \in S_{i+1}$, $W_i$ is the class with the highest index that contains worlds in which $\phi$ is true, and so $\Pi(\phi) = \sum_{w \in W_0, \ldots, W_i} f(w) = (r_1 - 0) + (r_2 - r_1) + \ldots + (r_{i+1} - r_i) = r_{i+1}$. □